\documentclass[conference]{IEEEtran}
\IEEEoverridecommandlockouts
\usepackage{cite}
\usepackage{amsmath,amssymb,amsfonts}
\usepackage{algorithmic}
\usepackage{graphicx}
\usepackage{textcomp}
\usepackage{xcolor}
\usepackage[flushleft]{threeparttable}
\usepackage{hyperref}
\def\BibTeX{{\rm B\kern-.05em{\sc i\kern-.025em b}\kern-.08em
    T\kern-.1667em\lower.7ex\hbox{E}\kern-.125emX}}
\usepackage[caption=false]{subfig}    
\begin{document}

\title{Single Image Haze Removal Using Conditional Wasserstein Generative Adversarial Networks}

\author{\IEEEauthorblockN{Joshua Peter Ebenezer, Bijaylaxmi Das, and Sudipta Mukhopadhyay\IEEEauthorrefmark{1}} \\
\IEEEauthorblockA{\textit{Dept. of Electronics and Electrical Communication Engineering} \\
\textit{Indian Institute of Technology Kharagpur}\\
Kharagpur, India \\
\IEEEauthorrefmark{1}smukho@ece.iitkgp.ac.in}}
\maketitle

\begin{abstract}
We present a method to restore a clear image from a haze-affected image using a Wasserstein generative adversarial network. As the problem is ill-conditioned, previous methods have required a prior on natural images or multiple images of the same scene. We train a generative adversarial network to learn the probability distribution of clear images conditioned on the haze-affected images using the Wasserstein loss function, using a gradient penalty to enforce the Lipschitz constraint. The method is data-adaptive, end-to-end, and requires no further processing or tuning of parameters. We also incorporate the use of a texture-based loss metric and the L1 loss to improve results, and show that our results are better than the current state-of-the-art.
\end{abstract}

\section{Introduction}
Fog and haze cause poor visibility and degrade the visual quality of images. Over the period of 2010-2016, an average of 25,451 crashes and 464 deaths occurred annually due to fog related accidents \cite{b1} in the US alone. Fog and haze can also affect the performance of computer vision algorithms for various tasks, particularly for automated driving. Haze is generally modeled using the Koschmeider law, which is a physics-based model that is given as follows

\begin{equation}\label{eq1}
I(x,y) = I_0(x,y)e^{-kd(x,y)} + I_\infty(1-e^{-kd(x,y)})
\end{equation}
\noindent where $I_0$ is the fog-free image, $x,y$ is the pixel location, $k$ is the fog extinction coefficient, $d(x,y)$ is the depth of the scene, $I_\infty$ is the sky intensity, and $I$ is the foggy image. The first term is called the attenuation term and the second is called the airlight map. 
\par
Haze-removal from a single image is thus an ill-posed problem, because it requires knowledge of the scene depth $d(x,y)$ as well the fog extinction coefficient $k$. A number of fog-removal methods~\cite{b2,b3} thus require multiple images of the same scene and calibrated cameras. On the other hand, single image fog-removal techniques rely on different priors in order to solve this under-constrained problem. He et al. proposed the dark channel prior~\cite{b4}, which assumes that the minimum intensity value across color channels in a patch is near to 0 in a clear, natural image. This prior has been widely used in conjunction with the Koschmeider law to obtain an initial estimate of the airlight map which is further refined by various methods and used to restore the image. The prior naturally fails when there are bright objects in the scene.
\begin{figure}
\centering
\includegraphics[width=.45\linewidth]{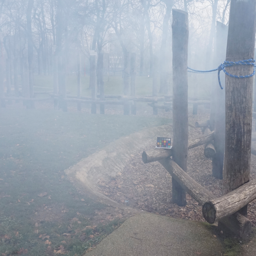}
\includegraphics[width=.45\linewidth]{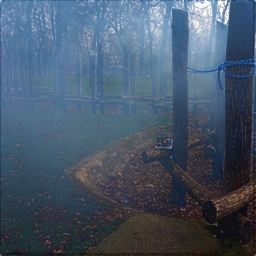}

\includegraphics[width=.45\linewidth]{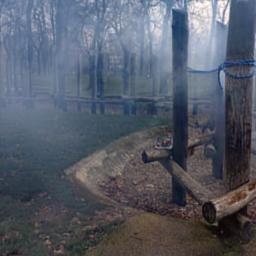}
\includegraphics[width=.45\linewidth]{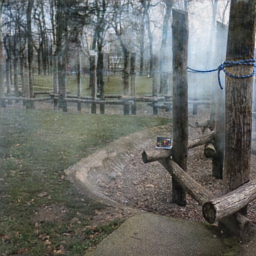}
\caption{ Clockwise from top left: Foggy image, Huang et al.'s method, Proposed, Li et al.'s method}
\end{figure}
We propose a deep learning based method that requires no prior on the image, and uses examples to learn the underlying relation between the haze-affected and the clear image. The method is end-to-end, and requires no parameters to be tuned at testing. A generative adversarial network (GAN)~\cite{b10} is used, with the Wasserstein penalty~\cite{b12} as the critic criteria and the use of perceptual texture loss as well as L1 loss. Unlike conventional GANs, Wasserstein GAN avoid the problems of vanishing gradients and mode collapse, and have better theoretical properties than the conventional loss functions for GANs. Training is stable and can generate high-quality images. These properties make WGANs ideal for image-to-image translation tasks such as haze-removal.  Our contributions are the use of the Conditional Wasserstein GAN (CWGAN) for image dehazing, and a new loss function that combines the Wasserstein loss, a perceptual loss, and the L1 regularization loss. We show our results are better than the state-of-the-art using widely accepted metrics, on datasets with and without reference images.
\section{Related work}

We briefly review recent work in the area of single image fog removal in this section, as well as recent research on conditional GANs and Wasserstein GANs. 

\subsection{Single image fog removal}

The dark channel prior (DCP) proposed by He et al.~\cite{b4} is derived from the assumption that the intensity value of at least one color channel within a local window is close to zero. Based on the DCP, the dehazing is done by estimating the airlight map, refining it, and then using it to restore the image using equation \ref{eq1}. Various modifications and improvements have been made on this assumption. Huang et al.~\cite{b5} incorporated the gray-world assumption into DCP to refine the estimate of the depth map. Zhu et al.~\cite{b6} proposed a model of the depth as a linear function of brightness and saturation, and learned the parameters by training. Broadly speaking, methods based on the DCP fail when there are bright objects in the scene, and generally require the user to tune a number of parameters for best results.
\par
Convolutional neural networks (CNN) have achieved great success at object recognition and classification tasks, and have also been used in fog removal applications. Cai et al. \cite{b8} proposed DehazeNet, a CNN that takes a hazy image, generates a transmission map, and restores the clear image using equation \eqref{eq1}. This method is sub-optimal because it does not allow the network to refine its estimates of the depth and the output implicitly by training in an end-to-end fashion. Li et al. \cite{b9} proposed an all-in-one network (AOD-net) that learns the mapping from a foggy to a clear image in an end-to-end manner. In the present work, we do not use a CNN with a handcrafted loss function. Instead we use a Wasserstein GAN that learns the conditional probability distribution in an adversarial manner, as the discriminator learns to distinguish between images produced by the network and the ground truth.

\subsection{Conditional Generative Adversarial Networks}

Goodfellow et al.~\cite{b10} proposed the generative adversarial network (GAN) to generate images (or text) from random noise samples. GANs consist of a generator and a discriminator. The generator tries to learn the probability distribution of the training samples and generate samples that can fool the discriminator into thinking they came from the training set. The discriminator tries to correctly identify samples as whether they come from the generator or from the training set. This is similar to a cop and counterfeiter game, where the counterfeiter tries to pass off counterfeited notes as real, while the cop tries to identify whether the notes he is shown are real or not. As the cop (the discriminator) and the counterfeiter (the generator) compete with each other, both get better at their tasks.
\par
GANs are difficult to train and face problems such as mode collapse and instability while training. Conditioning the output on some prior improves the training process. In a conditional GAN used for fog removal, the objective function is

\begin{small}
\begin{equation}\label{eq2}
\min_G \max_D \mathop{\mathbb{E}}_{I,G(I)\sim{\mathbb{P}_g}} [\log(1-D(I,G(I))] + \mathop{\mathbb{E}}_{I,J\sim{\mathbb{P}_r}} [\log(D(I,J))] 
\end{equation}
\end{small}
 
\noindent where $I$ is a prior input (foggy) image, $J$ is the clear (fog-free) image from the ground truth, and $G(I)$ is the output (fog-free) image produced by the generator $G$ when fed the prior image $I$. $G$ is called the generator and learns to imitate ${\mathbb{P}_r} $, which is the true joint distribution of the $I,J$ pairs of training examples, by generating $G(I)$ according to the joint probability distribution ${\mathbb{P}_g}$ conditioned on $I$. ${\mathbb{P}_g}$ is implicitly defined by $I, G(I)$ and is optimized by $G$ to mimic ${\mathbb{P}_r} $ exactly. Since $I$ is known to $G$ as a prior input, $G$ effectively learns the conditional probability of $J$ given $I$. $D$ is a CNN trained to correctly identify samples as whether they come from the true distribution ($J$ from ${\mathbb{P}_r} $)  or from the distribution that $G$ generates ($G(I)$ from ${\mathbb{P}_g}$), while having $I$ available as a prior.. In other words, it must assign the correct label to both training examples ($J$) and samples from $G$ ($G(I)$), similar to a policeman identifying currency notes as real or counterfeit A GAN without conditioning would not have $I$ available as an input prior, and would try to learn ${\mathbb{P}_r}$ from random noise and $J$.
 \par
Li et al.~\cite{b11} proposed a conditional generative adversarial network for haze removal. An image-to-image translation network with the architecture of a U-Net was trained to generate clear images from haze-affected images. A CNN was used as the discriminator. 
 \begin{figure*}%
    \centering
    \subfloat{{\includegraphics[width=.1\linewidth]{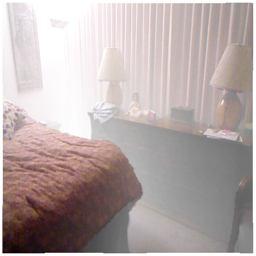} }}%
    \subfloat{{\includegraphics[width=.1\linewidth]{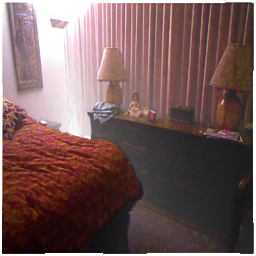}}}
    \subfloat{{\includegraphics[width=.1\linewidth]{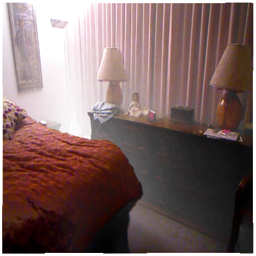}}}
    \subfloat{{\includegraphics[width=.1\linewidth]{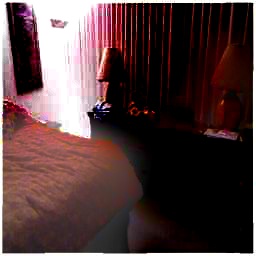}}}
    \subfloat{{\includegraphics[width=.1\linewidth]{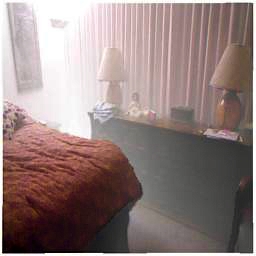}}}
    \subfloat{{\includegraphics[width=.1\linewidth]{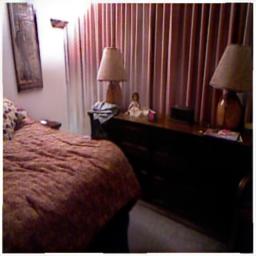}}}      
    \subfloat{{\includegraphics[width=.1\linewidth]{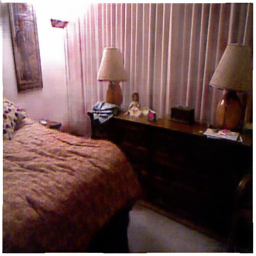}}}
                \subfloat{{\includegraphics[width=.1\linewidth]{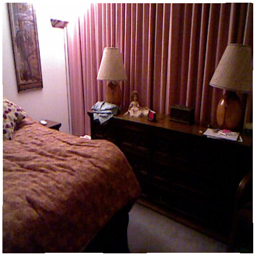}}} 
    \\
    \subfloat{{\includegraphics[width=.1\linewidth]{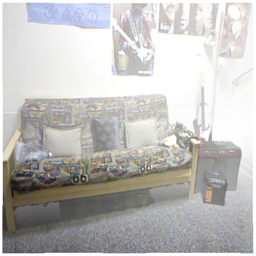} }}%
    \subfloat{{\includegraphics[width=.1\linewidth]{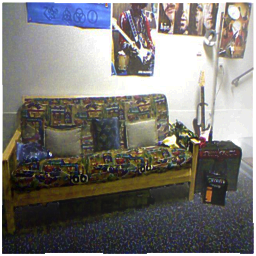}}}
    \subfloat{{\includegraphics[width=.1\linewidth]{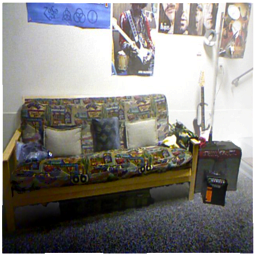}}}
    \subfloat{{\includegraphics[width=.1\linewidth]{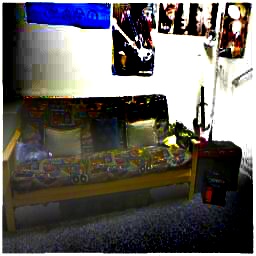}}}
    \subfloat{{\includegraphics[width=.1\linewidth]{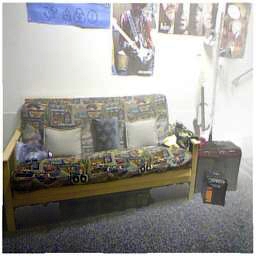}}}
    \subfloat{{\includegraphics[width=.1\linewidth]{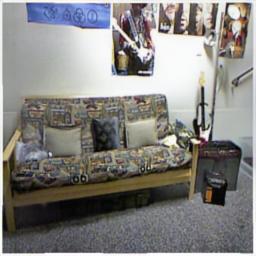}}}     
    \subfloat{{\includegraphics[width=.1\linewidth]{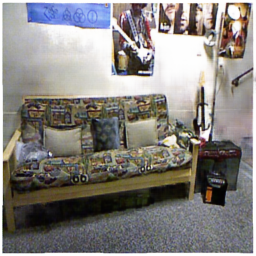}}} 
        \subfloat{{\includegraphics[width=.1\linewidth]{images/compete/synthetic/1202_real_B}}}    
    \caption{Results on synthetic images. From left to right: hazy image, Huang et al., Zhu et al., DehazeNet, AOD-net, Li et al., Proposed, Ground truth }%
    \label{fig:synthetic}%
\end{figure*}
\subsection{Wasserstein conditional Generative Adversarial networks}
Arjovksy et al.~\cite{b12} proposed the Wasserstein distance (or the Earth-mover (EM) distance) as an alternative loss function for GAN training. They showed that the EM distance could get rid of problems such as mode collapse and provide meaningful learning curves. Using the same notation as in the earlier section, the EM distance between two probability distributions $\mathbb{P}_r$ and $\mathbb{P}_g$ is defined as 

\begin{equation}\label{eq3}
W({\mathbb{P}}_r,{\mathbb{P}}_g) = \inf_{\gamma \in \prod({\mathbb{P}}_r,{\mathbb{P}}_g)}  \mathop{\mathbb{E}}_{(I,G(I),J)\sim{\gamma}} [||G(I)-J||]
\end{equation}

\noindent where $\prod({\mathbb{P}}_r,{\mathbb{P}}_g)$ denotes the set of all joint distributions $\gamma(I,G(I),J)$ such that $\mathbb{P}_g$ is the joint distribution of $I, G(I)$ and $\mathbb{P}_r$ is the joint distribution of $I,J$. Intuitively, the EM distance is the minimum cost required to transport the necessary mass from $G(I)$ to $J$ in order to transform $\mathbb{P}_g$ into $\mathbb{P}_r$ when conditioned on $I$. As finding this minimum cost is intractable, the Kantorovich-Rubinstein duality can be used to express the conditional WGAN two-player game as 

\begin{equation}\label{eq4}
 \min_G \max_{D\in{D_L}} \mathop{\mathbb{E}}_{I,J\sim{\mathbb{P}_r}} [D(I,J)] - \mathop{\mathbb{E}}_{I,G(I)\sim{\mathbb{P}_g}} [D(I,G(I))] 
\end{equation}

\noindent $D_L$  describes the 1-Lipschitz family of functions. In this formulation, $D$ is no longer called the discriminator, but is called the critic. This is because it does not perform classification, but simply produces a score that goes towards forming the objective function that is to be maximized with respect to $D$ and minimized with respect to $G$. Arjovksy et al. proposed to clip the weights of the critic in order to enforce the 1-Lipschitz constraint, which biases the critic towards simpler functions and requires careful tuning of the clipping parameter. Instead, Guljarani et al.~\cite{b13} proposed an alternative way to enforce the Lipschitz constraint, based on the observation that a function is 1-Lipschitz if and only if it has gradients of norm at most 1 everywhere. The new conditional WGAN objective then becomes:

\begin{equation}\label{eq5}
\begin{aligned}
 \min_G \max_{D\in{D_L}} \mathop{\mathbb{E}}_{I,J\sim{\mathbb{P}_r}} [D(I,J)] - \mathop{\mathbb{E}}_{I,G(I)\sim{\mathbb{P}_g}} [D(I,G(I))] \\ + \lambda \mathop{\mathbb{E}}_{I,\hat{J}\sim{\mathbb{P}_{\hat{j}}}} [||\nabla_{\hat{J}} D(I,\hat{J})||_2 -1)^2]
 \end{aligned}
\end{equation}
${\mathbb{P}_{\hat{j}}}$ is implicitly defined by $\hat{J}$ by sampling uniformly along straight lines between pairs of points sampled from the data distribution ${\mathbb{P}_r}$ and ${\mathbb{P}_g}$. In other words, $\hat{J}$ is defined by
\begin{equation}\label{eq6}
\hat{J} = \alpha J + (1-\alpha) G(I)
\end{equation}
\noindent where $\alpha$ is a linear interpolating factor and is randomly chosen. The motivation for this is that it can be shown that the optimal critic contains straight lines with gradient norm 1 connecting coupled points from ${\mathbb{P}_r}$ and ${\mathbb{P}_g}$. It is intractable to enforce the unit gradient norm everywhere and in practice this gives good results. This objective function ensures that the critic can be trained to optimality without the problem of vanishing gradients. The Jenson-Shannon divergence used to construct the objective function in \eqref{eq2} does not have this property, since the gradient vanishes as the critic approaches optimality. 
\section{Proposed method}

In this section we introduce the generator and critic architectures that we use. We also introduce a new loss function that combines the WGAN loss with other losses in order to improve the quality of the generated image.

\subsection{Network architecture}

The Generator is required to generate a clear, haze-free image from a hazy image. We use a U-Net~\cite{b145} as the generator. The U-Net accepts a haze-affected image as the input and is trained to generate the clear image. Information is compressed along one arm via repeated convolutions and max-pooling operations to a flattened vector at the bottom of the U. On the other arm of the U, transposed convolutions increase the dimensions of the information in a decoding process. Information generated by feature maps from the first arm is concatenated with the up-sampled feature maps of the second arm via skip connections to allow high-scale information to bypass the information bottleneck at the bottom. The penalized training objective of WGAN is not valid when batch-normalization is used, as the penalization is done with respect to each input independently and not the whole batch. Batch-normalization layers are thus not used in the generator.
\par
The critic is required to generate a high score for $I, J$ pairs of samples from the training examples, and to generate a low score for $I, G(I)$ pairs from the generator. Hence a convolutional neural network is used that takes two images as input, concatenates them to form a single input, and generates a score after appropriate convolution and max-pooling operations. The two images that are fed as input are the $I, J$ or the $I, G(I)$ pairs. The generator and critic architectures are the same as those used in the pix2pix network, and are described in detail in that work~\cite{b14}.
  \begin{figure*}%
    \centering
    \subfloat{{\includegraphics[width=.1\linewidth]{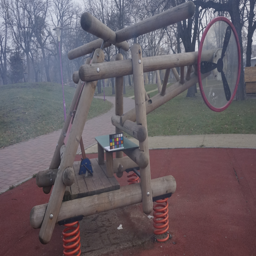} }}%
    \subfloat{{\includegraphics[width=.1\linewidth]{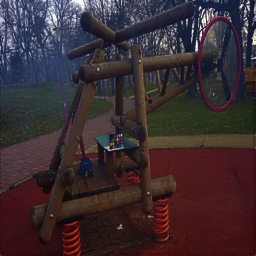}}}
    \subfloat{{\includegraphics[width=.1\linewidth]{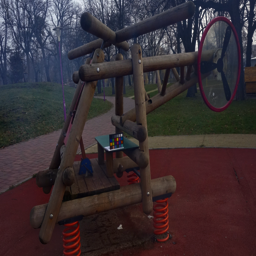}}}
    \subfloat{{\includegraphics[width=.1\linewidth]{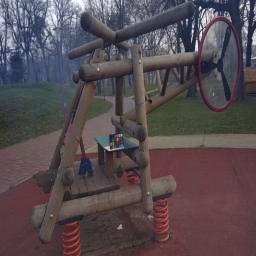}}}
    \subfloat{{\includegraphics[width=.1\linewidth]{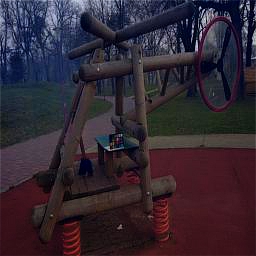}}}
    \subfloat{{\includegraphics[width=.1\linewidth]{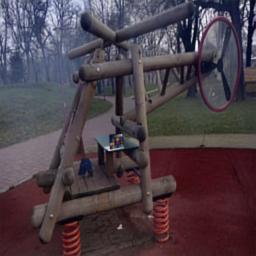}}}    
    \subfloat{{\includegraphics[width=.1\linewidth]{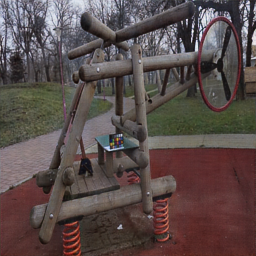}}}
                \subfloat{{\includegraphics[width=.1\linewidth]{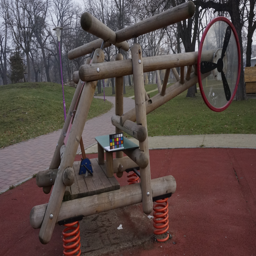}}} 
    \\
    \subfloat{{\includegraphics[width=.1\linewidth]{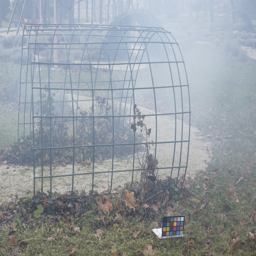} }}%
    \subfloat{{\includegraphics[width=.1\linewidth]{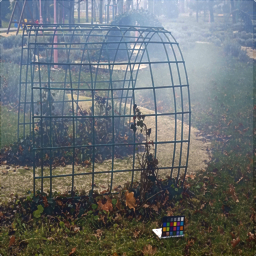}}}
    \subfloat{{\includegraphics[width=.1\linewidth]{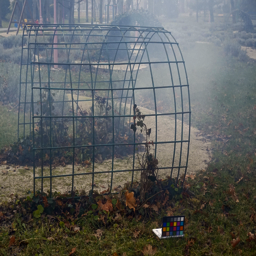}}}
    \subfloat{{\includegraphics[width=.1\linewidth]{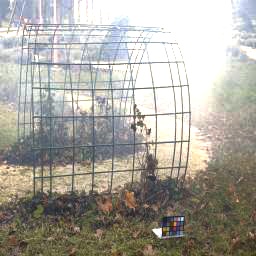}}}
    \subfloat{{\includegraphics[width=.1\linewidth]{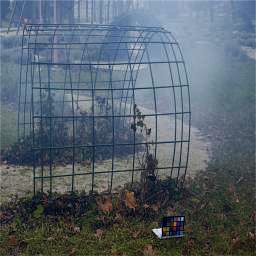}}}
    \subfloat{{\includegraphics[width=.1\linewidth]{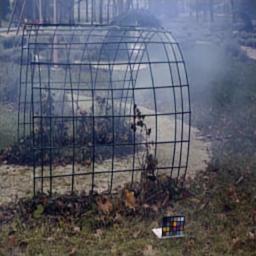}}}        
    \subfloat{{\includegraphics[width=.1\linewidth]{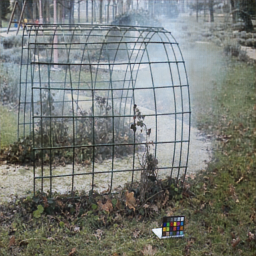}}}
            \subfloat{{\includegraphics[width=.1\linewidth]{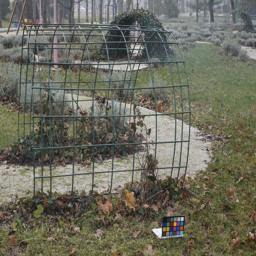}}}  
    \caption{Results on natural images. From left to right: hazy image, Huang et al., Zhu et al., DehazeNet, AOD-net, Li et al., Proposed, Ground truth }%
    \label{fig:natural}%
\end{figure*}
\subsection{Loss function}
 We use the VGG loss defined by Ledig et al.~\cite{b15}, that was shown to improve the perceptually relevant characteristics of the generated images. The VGG loss is defined by the ReLU activation layers of the VGG-19~\cite{b16} network, pre-trained on ImageNet. If the feature map corresponding to the 11\textsuperscript{th} layer of VGG when VGG is fed an input $I$ is denoted by $\phi(I)$, then the VGG loss for the conditional WGAN is defined as 
 
\begin{equation}
l^{VGG} =  ||\phi(J)-\phi(G(I))||^2_2
\end{equation} 
\noindent where $J$ is the reference clear image corresponding to the haze-affected image $I$ and $G(I)$ is the generator output, and the $L2$ norm is taken with respect to all the pixel values of the feature maps. We also use the $L1$ pixel-wise loss function, defined as

 \begin{equation}
l^{L1} =  ||J-G(I)||_1
\end{equation} 

\noindent which essentially enforces a $L1$ regularization prior. Combining these loss functions with the WGAN objective function, the generator is trained to minimize

\begin{equation}
L_G = \lambda_1 l^{VGG} + \lambda_2 l^{L1}-\mathop{\mathbb{E}}_{I,G(I)\sim{\mathbb{P}_g}} [D(I,G(I))]
\end{equation}

On the other hand, the critic is trained to maximize the following objective function.

\begin{equation*}
\begin{aligned}
L_D = \mathop{\mathbb{E}}_{I,J\sim{\mathbb{P}_r}} [D(I,J)]-\mathop{\mathbb{E}}_{I,G(I)\sim{\mathbb{P}_g}} [D(I,G(I))] \\ + \lambda_3 \mathop{\mathbb{E}}_{I,\hat{J}\sim{\mathbb{P}_{\hat{j}}}} [||\nabla_{\hat{J}} D(I,\hat{J})||_2 -1)^2]
\end{aligned}
\end{equation*}
 \section{Experiments and results}
 In this section, we present details of the experimental set-up as well as comparisons with various other competing methods.
\subsection{Dataset details}

\begin{figure*}[ht]
    \centering
        \subfloat{{\includegraphics[width=.1\linewidth]{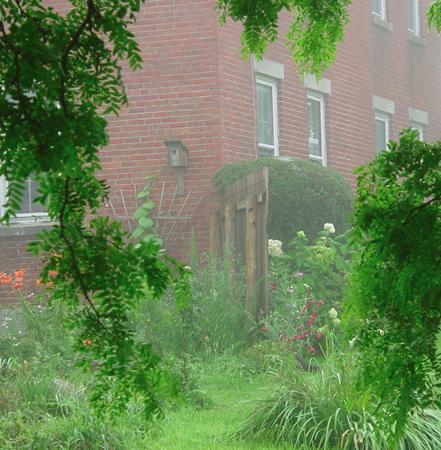} }}%
    \subfloat{{\includegraphics[width=.1\linewidth]{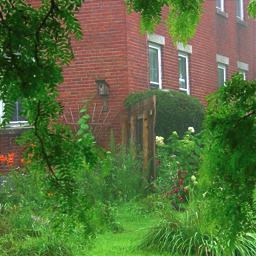} }}
    \subfloat{{\includegraphics[width=.1\linewidth]{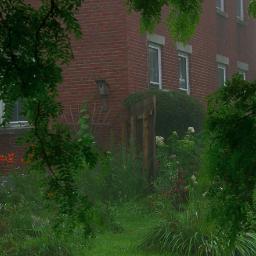} }}
    \subfloat{{\includegraphics[width=.1\linewidth]{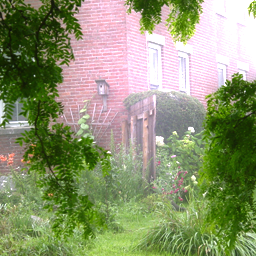} }}
    \subfloat{{\includegraphics[width=.1\linewidth]{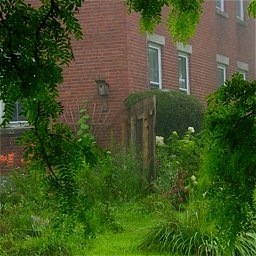} }}
            \subfloat{{\includegraphics[width=.1\linewidth]{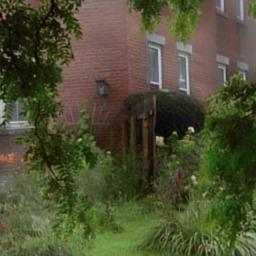} }}
        \subfloat{{\includegraphics[width=.1\linewidth]{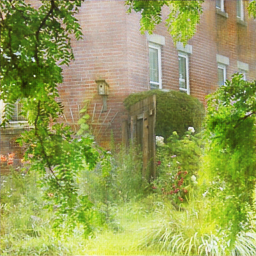} }}
    \\
    \subfloat{{\includegraphics[width=.1\linewidth]{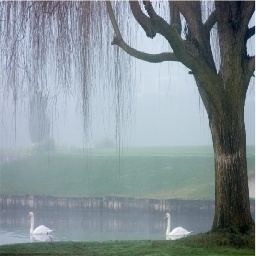} }}%
    \subfloat{{\includegraphics[width=.1\linewidth]{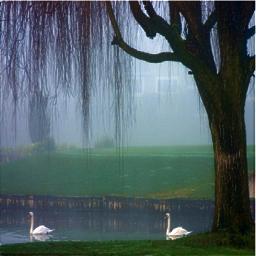} }}
    \subfloat{{\includegraphics[width=.1\linewidth]{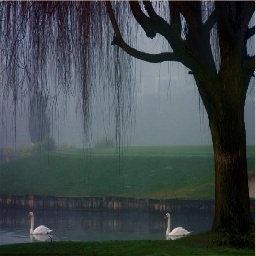} }}
    \subfloat{{\includegraphics[width=.1\linewidth]{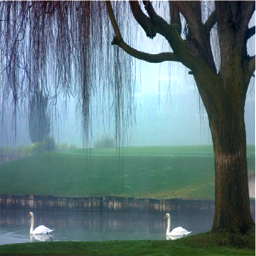} }}
    \subfloat{{\includegraphics[width=.1\linewidth]{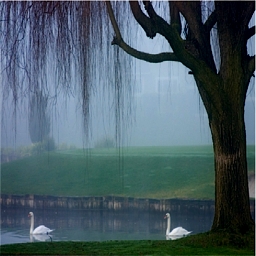} }}
            \subfloat{{\includegraphics[width=.1\linewidth]{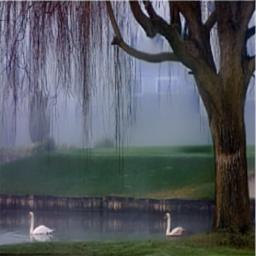} }}    
    \subfloat{{\includegraphics[width=.1\linewidth]{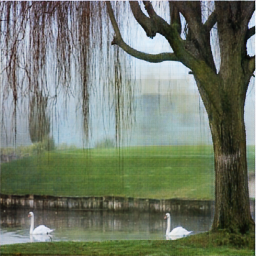} }}
    \caption{Results on natural reference-free images. From left to right: hazy image, Huang et al., Zhu et al., DehazeNet, AOD-net, Li et al., CWGAN }%
    \label{fig:noref}%
\end{figure*}

We use the publicly available D-Hazy~\cite{b17} and O-Haze~\cite{b18} datasets. The D-Hazy dataset consists of 1449 images of indoor scenes with synthesized fog created using \eqref{eq1} and the corresponding haze-free ground truths. The O-Haze dataset consists of 45 outdoor scenes (with haze-free ground truths) with haze generated by a professional haze machine that imitates real hazy conditions with high fidelity. An 80:20 split was made on both D-Hazy and O-Haze for training and test data. We also tested the model with a few images that do not have any reference image. The model was not trained on this set but produces commendable results. Results are shown in Fig.~\ref{fig:noref}.

\subsection{Experimental settings}

The input images and ground truth are resized to $256\times256\times3$ before they are fed to the WGAN. The size of the generator output is the same as that of the input. We use the Adam optimizer~\cite{b19} with a learning rate of $2\times10^{-4}$ and exponential decay rates $\beta_1=0.5$ and $\beta_2=0.999$. The network was trained for 1000 epochs on the training split of the D-Hazy dataset, and transfer-learned on the training split of the O-Haze dataset for 100 epochs. The critic is trained for five iterations for each iteration the generator is trained, and they are trained alternately. The networks were implemented on PyTorch and were run on a GeForce GTX 1080 Ti. The code and parameters are publicly available at \url{https://github.com/JoshuaEbenezer/cwgan}.

\subsection{Quantitative metrics}
We use the Peak Signal-to-Noise Ratio (PSNR) and Structural Similarity measure (SSIM)~\cite{b20} to compare the generator output with the reference images. We also use a few standard non-reference metrics, which are the gradient ratio ($r$), percentage of saturated pixels ($\sigma$), and contrast gain ($C$), to compare the generator output with the input foggy image~\cite{b21}. $r$ is the geometric mean of the ratios of the visible gradients in the output image to the foggy image. $\sigma$ is the percentage of pixels that are saturated in the output image but were not in the foggy image. $C$ measures the gain in contrast of the output with respect to the input. The higher the values of $r$ and $C$ and the lower the values of $\sigma$, the better the performance. Results are shown in Table~\ref{tab1}. Our method outperforms other methods on almost all measures. The high contrast gain obtained by AOD-net and Dehaze-net seem to be because images get over-saturated, as can be seen by the high value of the saturation percentage $\sigma$ for these two methods. On the other hand, the low value of $\sigma$ for Huang et al.'s method is because the contrast of the output image is not significantly greater than that of the input image, leading to a low value of $\sigma$ (which is desirable) but a low value of contrast gain $C$ (which is undesirable). 
\begin{table*}[h]
  \begin{threeparttable}
\caption[]{Quantitative metrics (Mean $\pm$ Std. dev.)\textsuperscript{1}}
  \centering
  \begin{tabular}{|c|c|c|c|c|c|c|c|}
\hline
Dataset & Metric & Huang et al.~\cite{b5} & Zhu et al.\cite{b6} & AOD-net~\cite{b9} & Dehaze-net~\cite{b8} & Li et al.~\cite{b11} & Proposed \\
\cline{1-8} 
D-Hazy & PSNR & 28.524$\pm$0.377 & 28.614$\pm$0.452 & 28.507$\pm$0.380 & 28.358$\pm$0.287&  29.435$\pm$0.836 & \textbf{29.941$\pm$0.807}\\
290 images & SSIM & 0.709$\pm$0.073 & 0.719$\pm$0.084 & 0.693$\pm$0.103 & 0.548$\pm$0.117 &  0.866$\pm$0.040 & \textbf{0.881$\pm$0.039} \\
(test split) & $r$ & 2.222$\pm$0.381 & 1.610$\pm$0.278  & 1.499$\pm$0.113 & 2.631$\pm$1.003 &  2.511$\pm$0.618 & \textbf{2.836$\pm$0.770} \\
 & $\sigma$ & \textbf{0.0690$\pm$0.2770} & 2.2886$\pm$3.5641   & 0.7057$\pm$1.2145 & 10.2029$\pm$9.3162   & 1.065$\pm$2.6938 &  0.6461$\pm$1.7447 \\
 & $C$ & 0.0719$\pm$0.0185 & 0.0805$\pm$0.0347 & 0.0532$\pm$0.0207 & \textbf{0.1798$\pm$0.0874} & 0.1101$\pm$0.0416 &  0.1075$\pm$0.0378\\
 \hline
O-Haze & PSNR & 27.849$\pm$0.441 & 27.925$\pm$0.276 & 27.810$\pm$0.142 & 28.371$\pm$ 0.441 & 28.215$\pm$0.320 & \textbf{28.859$\pm$0.887} \\
9 images & SSIM & 0.642$\pm$0.100 & 0.685$\pm$0.075 & 0.573$\pm$0.101 & 0.658$\pm$0.138& 0.687$\pm$0.106 & \textbf{0.848$\pm$0.848} \\
(test split) & $r$ & 1.692$\pm$0.413 & 1.508$\pm$0.423 & 1.720$\pm$0.265 & 2.132$\pm$0.615 & 2.415$\pm$0.845 & \textbf{2.947$\pm$0.813} \\
 & $\sigma$ & 0$\pm$0 & 0.4513$\pm$1.0533 & 0.6388$\pm$1.2794 & 0$\pm$0 & 0.0022$\pm$0.0028 & \textbf{0$\pm$0} \\
 & $C$ & 0.0653$\pm$0.0403 &  0.0563$\pm$0.0428 & \textbf{0.0850$\pm$0.0634} & 0.0234$\pm$0.009 & 0.0626$\pm$0.0212 & 0.0737$\pm$0.0114	 \\
 \hline
  \end{tabular}
  \begin{tablenotes}
  \item \textsuperscript{1}\footnotesize {Best values of each metric are marked in bold}
  \end{tablenotes}
  \label{tab1}
    \end{threeparttable}
\end{table*}
 \section{Conclusion}
We have proposed, for the first time, a conditional Wasserstein GAN for the image-to-image translation task of single image haze removal. We also introduce a new loss function that combines the Wasserstein loss with perceptual and regularization losses. The CWGAN achieves state-of-the-art results on all metrics for both datasets that were tested. The O-Haze dataset is markedly different from the indoor dataset, but the model is able to generalize well on O-Haze as well as reference-free images, indicating that it has learnt the underlying concept well.
Two major advantages of WGANs over conventional GANs is that they avoid the problems of vanishing gradients and mode collapse. Training is stable and the loss is well interpreted with respect to image quality. These properties make WGANs ideal for image-to-image translation tasks such as haze-removal. A future direction of work could be an experimental study of these properties in the context of image-to-image translation tasks such as haze-removal.

\bibliographystyle{IEEEtran}

\bibliography{IEEEabrv,conference_041818.bib}

\end{document}